\RequirePackage{fix-cm}
\documentclass[a4paper, 11pt]{article}

\usepackage{graphicx}
\usepackage{amsmath}
\usepackage{tikz}

\usepackage{float}
\usepackage{subcaption}
\usepackage{cite}
\usepackage{hyperref}
\usepackage{textcomp}

\usepackage{xcolor}

\usepackage[normalem]{ulem}

\newsavebox\CBox
\def\textBF#1{\sbox\CBox{#1}\resizebox{\wd\CBox}{\ht\CBox}{\textbf{#1}}}

\usepackage{pgfplots}
\pgfplotsset{width=7cm,compat=1.8}
\AtBeginDocument{%
  \providecommand\BibTeX{{%
    \normalfont B\kern-0.5em{\scshape i\kern-0.25em b}\kern-0.8em\TeX}}}

\DeclareCaptionFormat{custom}
{%
    \textbf{#1#2}\textit{\small #3}
}
\providecommand{\keywords}[1]
{
  \small	
  \textbf{\newline\textit{Keywords---}} #1
}

\begin{document}

\title{Audio-video fusion strategies for active speaker detection in meetings
}


\author{Lionel Pibre$^{1*}$       \and
        Francisco Madrigal$^{2}$ \and
        Cyrille Equoy$^{2}$ \and
        Fr\'{e}d\'{e}ric Lerasle$^{2}$ \and
        Thomas Pellegrini$^{1}$ \and
        Julien Pinquier$^{1}$ \and
        Isabelle Ferran\'{e}$^{1}$ \vspace{0,5cm}\and 
        $^{1}$ IRIT, Université de Toulouse, CNRS, INP Toulouse, UT3, Toulouse, France \\
 	    $^{2}$ LAAS-CNRS, UT3, Toulouse, France \\
 	    $^{*}$ Corresponding author: \texttt{lionel.pibre@gmail.com}
}




\date{}

\maketitle

\begin{abstract}

Meetings are a common activity in professional contexts, and it remains challenging to endow vocal assistants with advanced functionalities to facilitate meeting management. 
In this context, a task like \textit{active speaker detection} can provide useful insights to model interaction between meeting participants. Detection of the active speaker can be performed using only video based on the movements of the participants of a meeting. 
Depending on the assistant design and each participant position regarding the device, active speaker detection can benefit from information coming from visual and audio modalities. 
Motivated by our application context related to advanced meeting assistant, we want to combine audio and visual information to achieve the best possible performance. 
In this paper, we propose two different types of fusion (naive fusion and attention-based fusion) for the detection of the active speaker, combining two visual modalities and an audio modality through neural networks. 
In addition, the audio modality is mainly processed using neural networks. For comparison purpose, classical unsupervised approaches for audio feature extraction are also used. 
We expect visual data centered on the face of each participant to be very appropriate for detecting voice activity, based on the detection of lip and facial gestures. Thus, our baseline system uses visual data (video) and we chose a 3D Convolutional Neural Network (CNN) architecture, which is effective for simultaneously encoding appearance and movement. 
To improve this system, we supplemented the visual information by processing the audio stream with a CNN or an unsupervised speaker diarization system. We have further improved this system by adding visual modality information using motion through optical flow. 
We evaluated our proposal with a public and state-of-the-art benchmark: the AMI corpus. We analysed the contribution of each system to the merger carried out in order to determine if a given participant is currently speaking. 
We also discussed the results we obtained. Besides, we have shown that, for our application context, adding motion information greatly improves performance. Finally, we have shown that attention-based fusion improves performance while reducing the standard deviation.

\end{abstract}

\section{Introduction}
\label{sect:introduction}

Meetings are essential in our society, at universities and industries, where they are commonly held to coordinate professional matters such as projects, research, and funds, but can also be informal when people discuss everyday issues. Usually, at some point during a meeting, the person of interest is the one who is speaking because she/he is the center of attention of all the participants. 
This task of detecting the person(s) speaking at a given time is called active speaker detection. 
Active Speaker Detection is a multimodal analysis task that consists of analysing a video, determining if the movements of one of the faces appearing correspond to the speech signal contained in the audio track. This task can also be performed only on the video based on the movements of the participants.
Indeed, in such a context, communication takes place not only through voice but also relies on non-verbal signs as gestures, orientation, etc. 
Analysing this audiovisual information to estimate the person of interest or the active speaker can be useful in scenarios, such as human-robot interaction~\cite{He_2018} or human-machine multiparty dialogue~\cite{Haider_2016} among others, so a system able to interpret both cues is desirable. 
Various approaches to analyse meeting data have been carried out in the literature~\cite{Chakravarty_2016, Ren_2016}. 
Most of them focus on analysing audio-only, which indeed is a source that provides rich information. 
For example, multiple works explore ways to recognize the speaker~\cite{Vestman_2018}, others propose diarization techniques~\cite{Das_2017} that allow partitioning the incoming audio stream into homogeneous segments, i.e. audio sections with a single speaker label.
Likewise, some approaches process the audio, turning it into text~\cite{Yasir_2019}. As overlapping speech can lead to miss detections or recognition errors, audio detection may not be efficient enough, and the visual detection of the person may lead to some improvements.

Visual-only approaches for active speaker detection may encounter several difficulties. Indeed, for example, occlusions or facial movements such as facial expressions or yawns~\cite{Everingham_2006} can mislead these approaches. 
Nevertheless, the audio information combined with the video information helps to overcome these difficulties~\cite{Roth_2019}. Indeed, the movement of the lips and more generally of the head can help to detect if a person is speaking. 
Here, lips movement is encoded by spatiotemporal features, commonly estimated through Convolutional Neural Networks (CNN).

\section{Motivations}
\label{sect:Motivations}

This work was carried out as part of a project, and our objective is to analyse the audio and video data from the meetings in order to extract the maximum amount of information. 
For this work we do not want to analyse or process the personal information of the participants. The long-term objective is to bring out information that will be used for further analysis or processing, such as to perform segmentation in dialogue acts or to make a summary.

Here, we focus on active speaker detection during meetings based on video and audio cues. Meetings are a complex field of application as they are not always structured and have multiparty interactions. 

\begin{figure}[!htb]
\centering
  \includegraphics[width=0.8\textwidth]{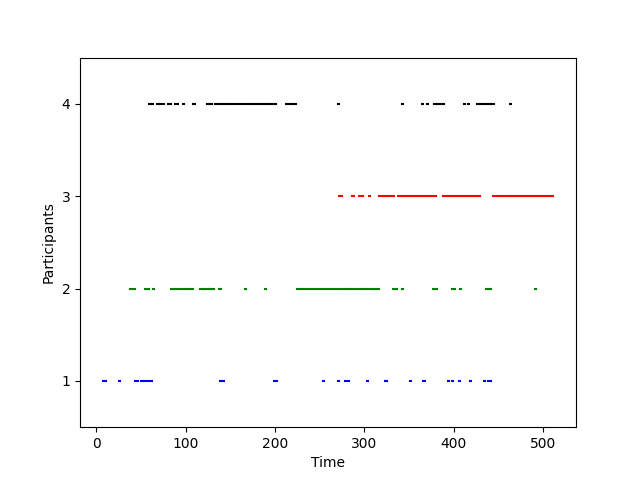}
  \caption{Illustration of the type of result we wish to obtain. The abscissa axis represents the time in seconds, and the ordinate axis represents the speakers. We want to detect each time a participant becomes active, even if another participant is already active.
  \label{fig:active_speaker}}
\end{figure}

\begin{figure}[!htb]
\centering
  \includegraphics[width=0.7\textwidth]{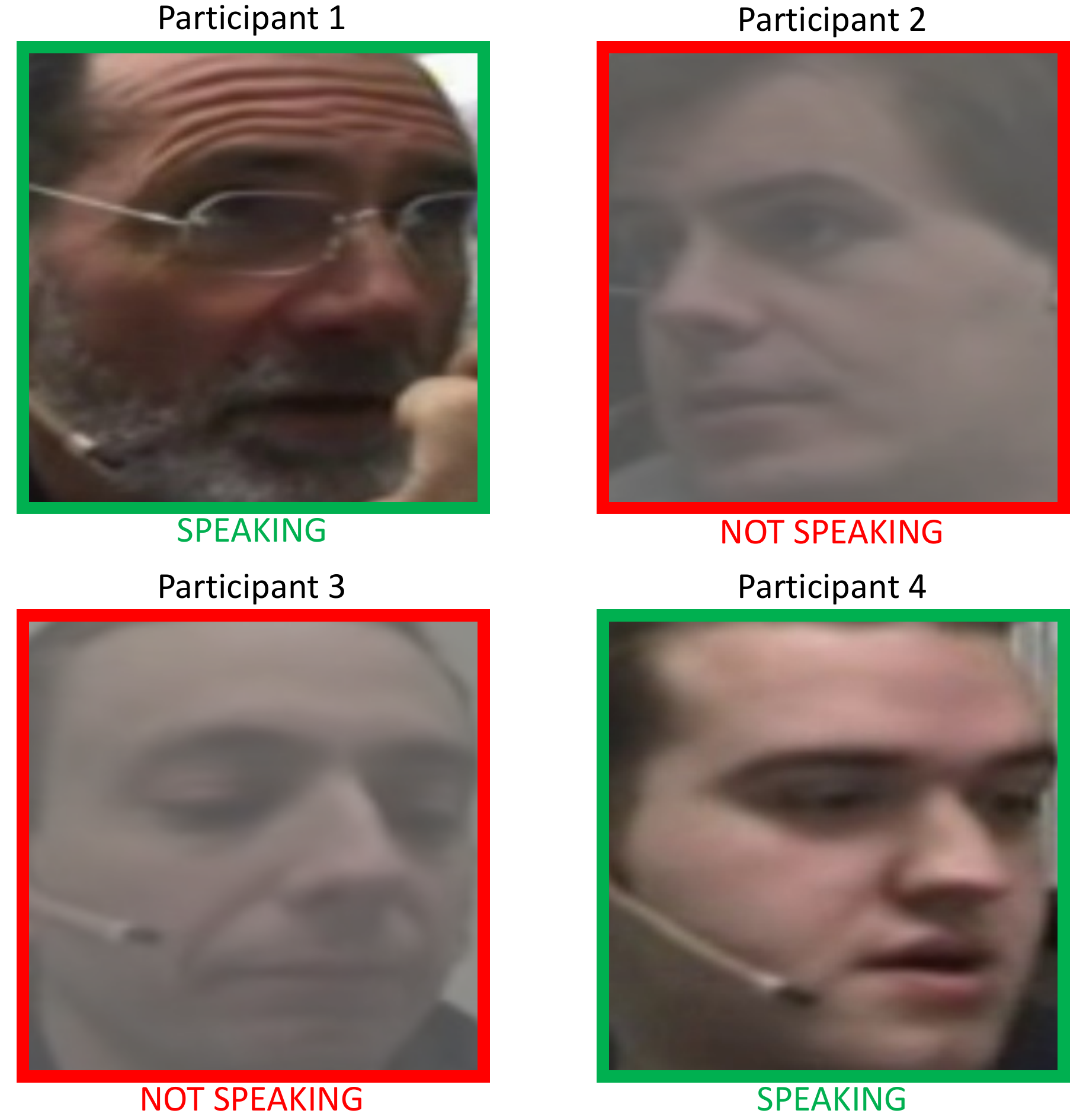}
  \caption{Illustration of the type of result we wish to obtain. For this example, we look at the result obtained at a given time. In this case, we can see that participants 1 and 4 are talking, and participants 2 and 3 are listening and are therefore not active speakers.
  \label{fig:active_speaker2}}
\end{figure}

For this issue, it is not only a question of determining whether someone is speaking (Voice Activity Detection) but also which participant is speaking. The figures~\ref{fig:active_speaker} and \ref{fig:active_speaker2} illustrate our problem. 

As can be seen in the two figures, in our problem it is quite possible that several speakers are active at the same time. 
For this goal, our proposal focuses on three aspects: 
\begin{itemize}
    \item merge visual and audio features such that we obtain a robust final detection,
    \item find the most suitable fusion for our problem,
    \item compare two possible approaches for the analysis of the audio modality, one based on neural networks and the other on an unsupervised method.
\end{itemize}

As previously explained, active speaker detection can be carried out either by using video alone or by combining the visual modality with the audio modality.
Our approach is based on the prior image detection of the face using state-of-the-art techniques. The principle is to analyse frames to characterize the orientation of the face in a continuous space in terms of pitch, yaw, and roll. Therefore, a "speaker" / "non-speaker" status can be inferred at once with only the pure visual percept.

However, in literature, there are few public multi-person audio-video datasets in our application context (meeting), which limits the scope of the evaluations and comparisons that we can perform. 
Nevertheless, the evaluations of the available datasets with increasing complexity, from mono to multi-person observed within the field of view, are encouraging for the different analyses and experiments that we have carried out.
The main contributions of this paper are:
\begin{enumerate}
    \item A pure visual speaker classifier, based on 3D CNNs, is applied in this original context. It takes as input video frames (i.e. clips) with the Red, Green, and Blue channels but also optical flow information.
    
    \item The comparison of two approaches for audio modality analysis: a supervised approach with a neural network, and an unsupervised approach with a speaker segmentation and clustering method.
    
    \item The comparison of two fusions: a fusion called \textit{naive} and a fusion based on \textit{attention modules}.
\end{enumerate}

This paper has the following structure: section~\ref{sect:related_work} presents the related work and section~\ref{sect:modalities} we explain the processing that we apply to each of the modalities.
Section~\ref{sect:fusion_strategies} describes the strategies used for the fusion. Section~\ref{sect:experiences} details the experimental protocol used, and the quantitative and qualitative results, including a discussion, are given in section~\ref{sect:evaluation}. Last, section~\ref{sect:conclusion} concludes and mentions future work.

\section{Related work}
\label{sect:related_work}

Communication situations and interaction, between humans commonly involve voice and gestures, encoding verbal and non-verbal signs, which computer process as audio and video signals respectively. 
Our work focuses on exploiting these signals to estimate the person (s) who is speaking at a given time, and point him/her then as the active speaker. From both signals, audio has been widely explored because it naturally captures whether somebody is speaking or not.

\subsection{Audio-based methods}

In literature, there is a large amount of work on speaker recognition~\cite{Vestman_2018}. For such purpose, one way to achieve it involves speaker diarization techniques~\cite{Das_2017} where the audio stream is partitioned into "speech" and "non-speech" segments, and where a speaker ID is assigned to each "speech" segment. 
In~\cite{Bonastre_2011}, Bonastre et al. proposed a speaker diarization method based on the binary key modeling, which transforms the audio into a feature representing the speaker within the binary space. Then, the diarization is performed by an iterative agglomerative clustering algorithm that forms segments of a same speaker. 
Patino et al.~\cite{Patino_2018} improved the method by considering spectral clusterization. One of the main challenges is to be able to recognize the same person regardless of the intensity of the speaker's voice, e.g. whispering, or the background noise which may alter speaker identification.
Vestman et al.~\cite{Vestman_2018} did a deep taxonomy on different features that address these issues and proposed a sound time-varying feature that gives state-of-the-art results. With the rise of the Convolutional Neural Network (CNN), some proposals~\cite{Hruz_2017} have exploited such end-to-end solution for speaker diarization. 
The most recurrent architectures are those based on Long Short Term Memory (LSTM) Networks (\cite{Sarkar_2018, Wang_2018}) since they capture the variations in the voice of the speaker. One of the limitations of CNNs is that they are computationally expensive, generally requiring powerful GPUs to produce good results. 
In~\cite{Xie_2019}, the authors propose an end-to-end system at utterance-level for speaker verification. This method proposes a new "\textit{thinResNet34}" trunk architecture, which incorporates a \textit{GhostVLAD} layer allowing to aggregate features across time. 
The \textit{thinResNet34} architecture has only 3 million parameters when a classic \textit{ResNet-34}~\cite{He_2016} has 22 million. It is trained and evaluated using the VoxCeleb dataset~\cite{Nagrani_2020}, an audiovisual dataset of short clips of interviews extracted from YouTube. 
In this dataset, with over 2000 hours of recordings and more than 7000 people, \cite{Xie_2019} has demonstrated the effectiveness of such a compact network by providing state-of-the-art performances.

However, whispers, background noises, or interspersed audio cause bad estimates. To overcome these difficulties, several approaches include visual features since those are not affected by these phenomena.

\subsection{Video-based methods} 

Pure visual information is an important source to consider speaker recognition, especially if the audio is not available, corrupted, or unintelligible. 
Zhou et al.~\cite{Zhou_2014} give an in-depth review of advances in visual speaker recognition until 2014. The authors provided a list of datasets aimed for this purpose. Also, the presented methods are grouped according to the type of features used, highlighting three groups: 
\begin{itemize}
    \item Image-based group. Here raw pixels are transformed directly as visual features with the aid of methods such as Principal Component Analysis (PCA)~\cite{Hong_2006, Le_2016}. 
    More recently, the authors of~\cite{Stefanov_2017} have used neural networks for the detection of active speakers. To do so, it first uses a face detector, then the authors use the AlexNet~\cite{Krizhevsky_2012} network to extract features on RGB color data for all found faces, and finally they give these features to a recurrent LSTM network. 
    
    \item Motion-based group. Features describe the observed movement during speaking. Instead of creating handcrafted features, several proposals estimate the motion directly from the video with techniques such as Optical Flow (OF)~\cite{Le_2016}. 
    
    \item Geometric-based group. The features focus on the geometric information of a moving mouth. In~\cite{Korshunov_2019}, the movement is computed by measuring the distance between points detected over the mouth. Then, the difference of distances between successive images represents the motion. 
    The main limitation of these methods is that any movement of the mouth is associated with a speaking status. So the number of false positives is commonly high. To overcome this problem, this motion information can be combined with the audio signal and/or the raw images.
\end{itemize}

\subsection{Audiovisual-based methods} 

Audiovisual methods are robust to background noise and different speaking modes (whisper). Recently deep learning has provided advances in audiovisual speaker estimation by capturing the temporal relationships of visual and acoustic cues. 
The use of a recurrent neural network (RNN) is explored in~\cite{Tao_2019} to extract video features using 2D CNN. Then, in a similar way as~\cite{Korshunov_2019}, they train an LSTM layer for each modality. The output of both layers is concatenated and the result is used to train a final LSTM layer. 
This process is known as early fusion, on the contrary, it is called late fusion when the outputs are merged at the end without any other training layer pursuing them. 
In~\cite{Petridis_2018}, Petridis et al. compares both fusion methods in the context of speaker recognition. Afouras et al.~\cite{Afouras_2019} explore the impact of training audiovisual lip-reading network with different loss functions.

In the literature, the application of LSTM layers has been widely studied because it is good at capturing spatiotemporal information related to a speaker. 
Besides, other CNN architectures learn this aspect but are used in different classification contexts. Such is the case of the deep 3-Dimensional Convolutional Network (C3D)~\cite{Tran_2015}; here the objective is to classify actions such as "\textit{biking}", "\textit{running}", among others.
This 3D CNN model has been extended to other architectures, in~\cite{Tran_2018} they study the use of a 3D residual neural network (\textit{ResNet3D}). 
In both cases, the goal is to obtain a 3D feature that encodes simultaneously appearance and motion. We aim to study these networks in our context of speaker detection because these proposals show good performances with state-of-the-art methods. 
The classic CNN-based image classification approach learns from raw images. Some works have shown that the use of additional features like optical flow~\cite{Le_2016, Borghi_2017, Tran_2018} or depth cue~\cite{Borghi_2017}, help to improve estimation.


\subsection{Synthesis} 
The main drawback of video-based applications (whether pure or in conjunction with audio) is that their evaluations are conducted in constrained situations where there is usually only one person, or few people looking straight at the camera. 
This scenario is not realistic in the context of a traditional meeting. Therefore, we propose to use a classifier, which considers spatiotemporal features, trained from videos in realistic meeting situations. 

Additionally, the video can provide more information if we consider the social aspect among the meeting participants. For example, a person who speaking will move, have facial expressions and make gestures to support his or her arguments. That is why in this work, different fusion strategies have been studied. First, let's describe how each modality has been processed.

\section{Mono-modality processing}
\label{sect:modalities}

\subsection{Visual modality processing} 

Let's start by looking at the visual modality. We worked with two visual modalities: the RGB images of the videos, and the optical flow image calculated from these images.

\subsubsection{Video-based network architecture}

In order to obtain a visual representation, we consider a 3D CNN architecture. Since we want to capture the temporal relationship that exists between the images, we group several consecutive images to form a non-overlapped video clip of $L$ frames. 
The input clip has a size $3\times L \times H \times W$, where $H$ and $W$ are the height and width of the frame in the Red, Green, and Blue (RGB) channels. We relied on~\cite{Tran_2015} to set this parameter and chose $L = 16$ as clip size, i.e. the number of frames. Thus, the clips are given as input to the networks. 

In comparison with 2D CNN, here we have a 4-dimensional tensor. The difference with a classic 2D CNN is that the convolution is performed in 3D, allowing the feature maps to learn temporal context. Thus, the convolution layers create a 3D feature where the initial part focuses on the appearance of the first images and the rest considers the salient motion~\cite{Tran_2015}.

In our case, we chose to use the \textit{ResNet3D} architecture. This architecture takes up the idea of \textit{ResNet} residual blocks~\cite{He_2016} but using a 3D convolution instead of a 2D one. 
This preserves and propagates the temporal reasoning through the network layers. Input is a 4D tensor and the image size being the same as \textit{ResNet2D}: $3 \times 16 \times 224 \times 224$. Here, we evaluate the \textit{ResNet3D-18} version, i.e. considering 18 blocks. In the rest of this paper, we will call this method "Video-net".
Video-net is trained with RGB images that show only one participant's face.

The output of this network indicates the probability that the analysed participant is speaking or not. 

\subsubsection{Optical flow calculation and processing}

In order to add the facial movement information, we took inspiration from the works~\cite{Le_2016, Borghi_2017, Tran_2018}, and encoded this information through the optical flow.

\begin{figure}[!htb]
\centering
\begin{subfigure}{\linewidth}
\includegraphics[width=\linewidth]{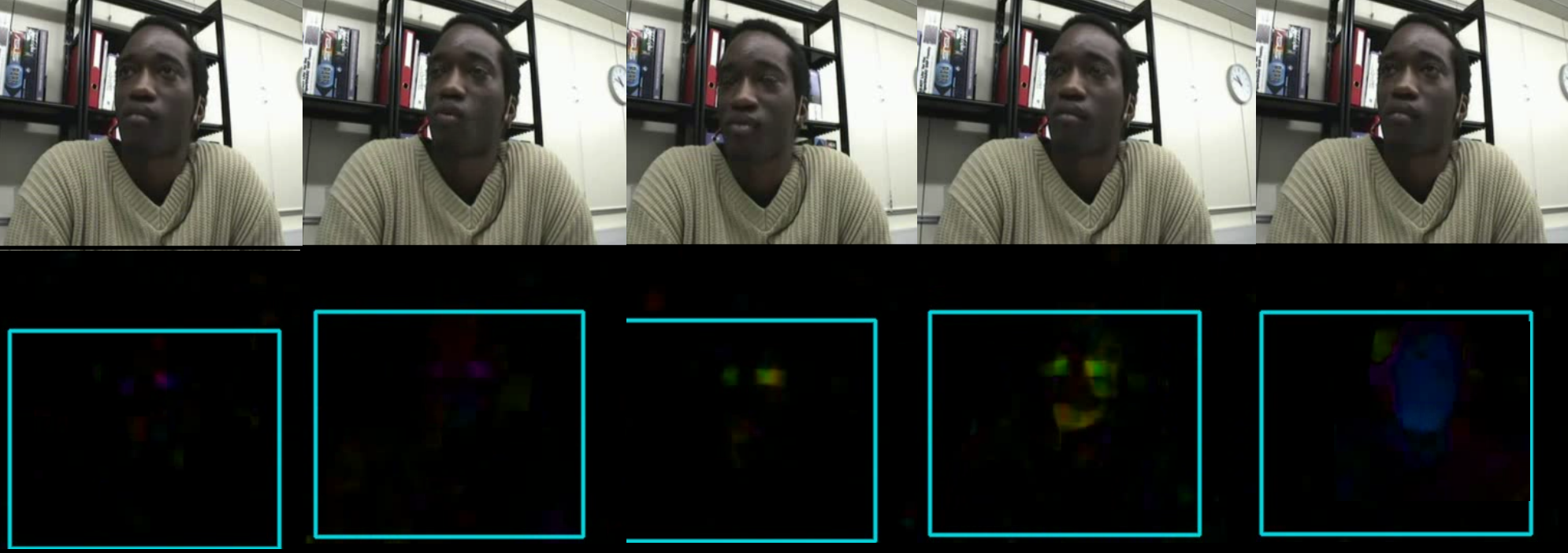}
\caption{}
\label{fig:blink}
\end{subfigure}\\
\vspace{0.5cm}

\begin{subfigure}{\linewidth}
\includegraphics[width=\linewidth]{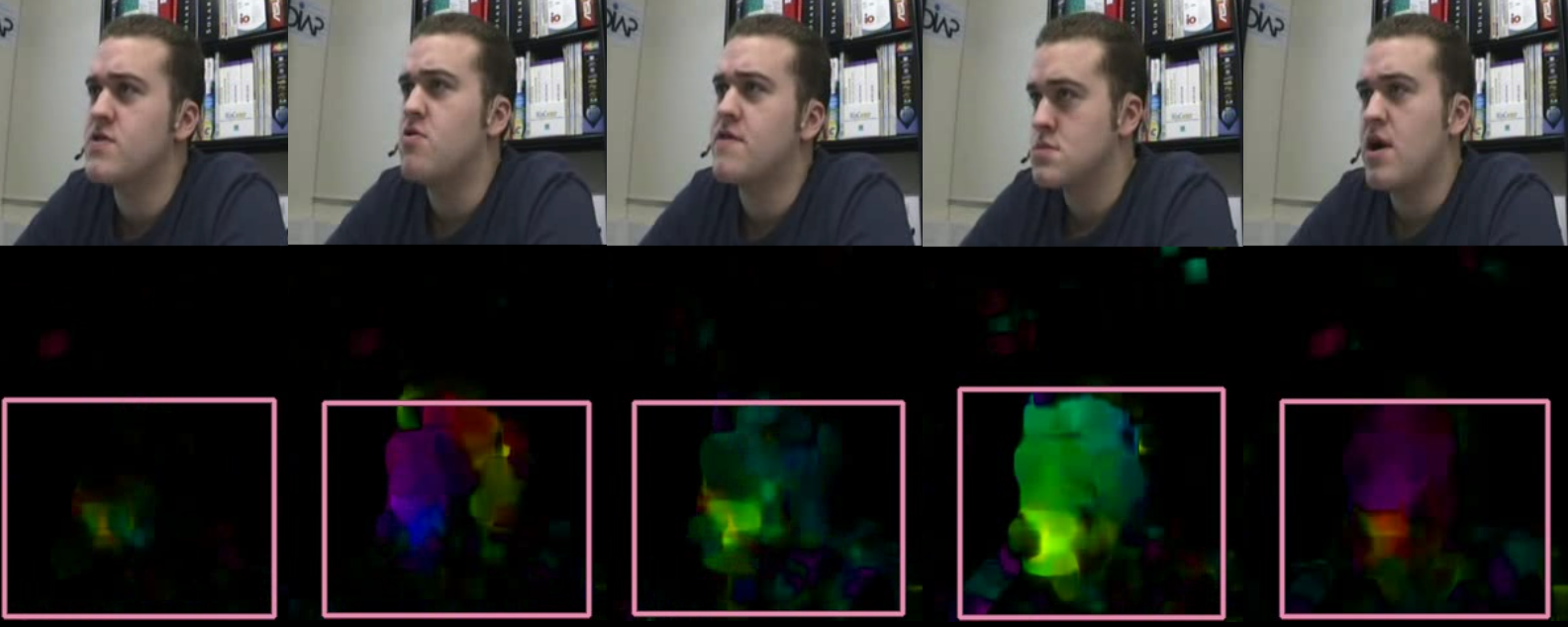}
\caption{}
\label{fig:speak}
\end{subfigure}

\caption{ We show in this figure two examples, (a) and (b), of images given to CNN. For these two examples, we have at the top the images of the original video and at the bottom the magnitude of optical flow images. We can clearly see that the person in example (a) is blinking while the person in example (b) is speaking.\label{fig:OF}}
\end{figure}

We present two examples in figure~\ref{fig:OF}, where we have for each example at the top the RGB images and at the bottom the magnitude of optical flow images. As you can see, the person in example (a) is blinking while the person in example (b) is speaking. 

We trained a network with two branches: one branch for the raw image and one for the optical flow. Each of the branches is a \textit{ResNet3D}. We remove the top of each network and concatenate both visual representations. Finally, we trained two fully connected layers and a softmax loss layer.


\subsection{Audio modality processing} 

Now let's focus on the two audio approaches we used.

\subsubsection{Audio-net system}

For the audio representation, we use the author’s implementation of~\cite{Xie_2019}, named VGG-Speaker-Recognition framework\footnote{\url{https://github.com/WeidiXie/VGG-Speaker-Recognition}}. 
The first part of this network (\textit{thinResNet34}) takes as input an extract of the spectrogram computed from the audio file. This network uses the principle of shared weights on all extracts belonging to the same audio file.

The second part of this network is made up of a dictionary-based \textit{GhostVLAD} layer~\cite{Zhong_2018} to aggregate features across time. The objective of this part is to obtain a vector of a fixed size as output of the network whatever the length of the audio file processed.

The spectrograms are calculated from audio clips corresponding to the duration of the video clip, i.e. 0.64 seconds, with 256 frequency components. The spectrogram is normalized by subtracting the mean and dividing by the standard deviation. The \textit{thinResNet34} network was trained with Adam optimizer and an initial learning rate of $1e$-$2$. 

This network produces an audio representation that is initially intended to respond to a speaker recognition task, or a speaker verification task. For our problem, the vector produced by this network will be merged with other vectors coming from other systems components applied to audio or video data. In the rest of this paper, we will call this method "Audio-net".

\subsubsection{pyBK speaker diarization system}

After carrying out a study on the existing speaker diarization system applied to meeting data from AMI, \textit{pyBK}~\cite{Patino_2018} was selected as showing the best performances. 
To be more specific, we used the output of \textit{pyBK}. Indeed, in the case of \textit{pyBK}, we directly merge the result of this approach with Video-Net and Audio-net presented earlier. 
To do this we used hot vectors. 
A hot vector is a representation of categorical variables as binary vectors. Here, we took as vector size the maximum number of speakers found by \textit{pyBK} applied on the whole set of meeting recordings. 
So we created hot vectors of a dimension $N_{max} + 1$, to be able to encode when no one is speaking. To merge this hot vector with Video-net and Audio-net, we use a Gated Recurrent Unit~\cite{Cho_2014} (GRU) layer proposed by Cho et al.~\cite{Cho_2014}. 
Its role is to make each recurrent unit to adaptively capture dependencies of different time scales. The GRU has gating units that modulate the flow of information inside the unit.

The activation $h_t^j$ of the GRU at time $t$ and layer $j$ is a linear interpolation between the previous activation $h_{t-1}^j$ and the candidate activation $\tilde{h}_t^j$:
\begin{equation}\label{eq:h1}
    h_t^j = \tilde{h}_{t}^j (1 - z_t^j) + z_t^j h_{t-1}^j
\end{equation}

where an update gate $z_t^j$ decides how much the unit updates its activation. 

The update gate is computed by:
\begin{equation}~\label{eq:h2}
    z_t^j = \sigma\left([\mathbf{W}_z \mathbf{x}_t]^j + [\mathbf{U}_z \mathbf{h}_{t-1}]^j\right)
\end{equation}

where $\sigma$ is a logistic sigmoid function, and $\mathbf{x_t}$ is an element of a given sequence $\mathbf{x} = (\mathbf{x_1},\mathbf{x_2},\dots,\mathbf{x_T})$. The logistic sigmoid function will transform the values between 0 and 1, allowing the gate to filter between the less-important and more-important information in the subsequent steps.

The candidate activation $\tilde{h}_t^j$ is computed as in equation~\ref{eq:h3} from~\cite{Bahdanau_2014}.

\begin{equation}\label{eq:h3}
    \tilde{h}_t^j = \tanh\left([\mathbf{W} \mathbf{x}_t]^j + [\mathbf{U}(\mathbf{r}_t \odot \mathbf{h}_{t-1})]^j\right)
\end{equation}

where $r_t$ is a set of reset gates, $\tanh$ is hyperbolic tangent, and $\odot$ is an element-wise multiplication. When off ($r_t^j$ close to 0), the reset gate effectively makes the unit act as if it is reading the first symbol of an input sequence, allowing it to forget the previously computed state.

The reset gate $r_t^j$ is computed similarly to the update gate:
\begin{equation}\label{eq:h4}
    r_t^j = \sigma\left([\mathbf{W}_{r} \mathbf{x}_t]^j + [\mathbf{U}_{r} \mathbf{h}_{t-1}]^j\right)
\end{equation}

Since the input of Video-net is clips of 16 frames, we have extracted the result vector of \textit{pyBK} at the time of each frame. Thus, we have at the input of the GRU layer 16 vectors of size 8.

\section{Fusion strategies}
\label{sect:fusion_strategies}


The previous steps have calculated vectors that represent if a participant is a speaker or not by analysing different cues. Our goal is to merge such feature vectors, two approaches are proposed.

\subsection{Naive fusion}

For this fusion, we have applied the easiest way to merge these vectors, by concatenating them. However, one of the vectors is likely to be more discriminating than others. This is why we made the size of the various vectors vary to be able to analyse the importance of each method on the fusion process.

\begin{figure}[!htb]
\centering
  \includegraphics[width=\textwidth]{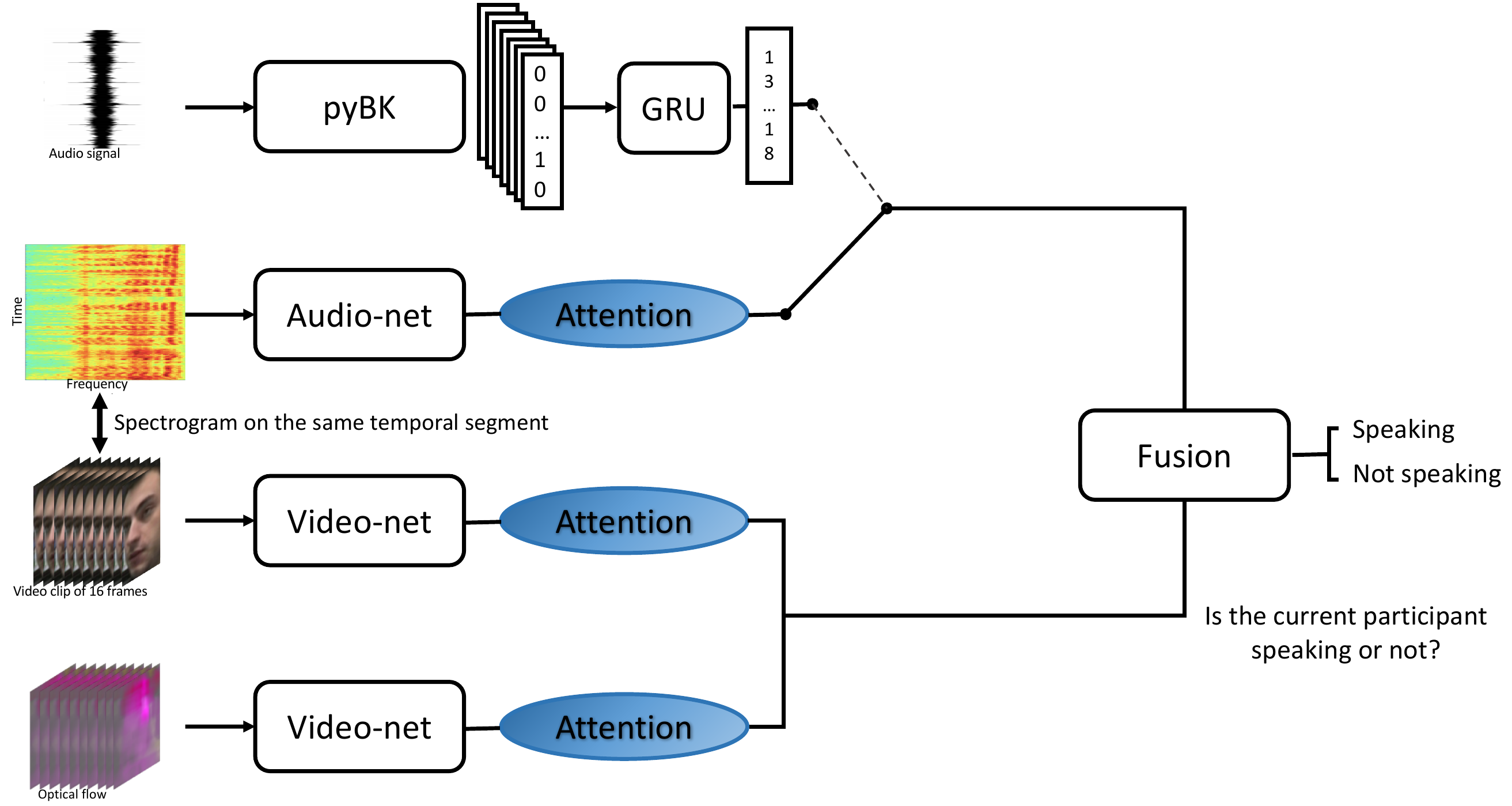}
  \caption{Architecture used for naive fusion. The top two branches are associated with audio modality processing. In this work, \textit{pyBK} is an alternative to Audio-net, so these two branches are never activated at the same time, either Audio-net or \textit{pyBK} is used. 
  The two lower branches are associated with visual modality processing via Video-net. Attention is not used in the naive fusion approach.
  \label{fig:fusion}}
\end{figure}

We present, in figure~\ref{fig:fusion}, the global architecture used to perform the fusion. In the case of naive fusion, attention modules (in blue) are not used. The two top branches are associated with audio modality processing. These two branches are never activated at the same time, either Audio-net or \textit{pyBK} is used. The two lower branches are associated with visual modality processing via Video-net.


\subsection{Attention-based fusion}

We then used a second fusion strategy based on attention layers.
The principle of this fusion is to apply an attention layer (as shown in figure.~\ref{fig:fusion} in blue) to each of the modalities and to merge these results.

The principle of the attention mechanism is to give more importance and to focus on the elements of a feature vector that are the most discriminating. In practice, attention is simply a vector, often the outputs of a dense layer using softmax function.

Let $\textbf{\^{x}}^m$ be the result of applying the softmax on the output vector $\textbf{x}^m$ of a modality $m$ with $\textbf{x}^m = x^m_1 \dots x^m_n$, where $n$ is the representation dimension. $\textbf{\^{x}}^m$ is computed as:
\begin{equation}\label{eq:attention1}
    \textbf{\^{x}}^m = \left(\frac{e^{x_i^m}}{\sum\limits_{j} e^{x_j^m}}\right)
\end{equation}

Finally, to obtain the output vector $\textbf{o}^m$ for this modality, we multiply term by term the vectors $\textbf{\^{x}}^m$ and $\textbf{x}^m$:
\begin{equation}
    \textbf{o}^m = \textbf{\^{x}}^m \cdot \textbf{x}^m
\end{equation}

These $\textbf{o}^m$ vectors are then concatenated.

\section{Experimental protocol}\label{sect:experiences}

Our experiments have been carried out using the Keras Python library on a workstation with an Intel Xeon E-2286G 4.0 GHz CPU, 64 Gb of RAM, and one NVIDIA GeForce RTX 2080 Ti. 
We also used the OSIRIM computing cluster\footnote{\url{https://osirim.irit.fr/site/en}} composed of 7 GPU servers each equipped with 2 Xeon 2640 V4 @ 2.40 GHz CPUs and 4 NVIDIA GeForce GTX 1080 Ti GPUs.

\subsection{Dataset}

We have evaluated our method using the AMI Corpus (Augmented Multi-party Interaction)~\cite{Carletta_2007}. This dataset consists of over 100 hours of meeting recordings. 
The recordings use a range of signals synchronized to a common timeline and recorded with different sensors: close-talking and far-field microphones, individual and room-view video cameras. It also includes outputs from a slide projector and an electronic whiteboard. During each meeting, the participants have also unsynchronized pens available to them that record what is written. 

The meetings were recorded in English using three different rooms with different acoustic properties, and include mostly non-native speakers. To make sure that we keep the same meeting context (microphones, cameras...), we have selected the IDIAP "remote control scenario" meetings. 
The information for this scenario is presented in the table~\ref{tab:info_IDIAP}. 

\begin{table}[htb]
\small
\centering
\caption{Information about the IDIAP remote control scenario meetings.
\label{tab:info_IDIAP}}
\begin{tabular}{cccc}
\hline\noalign{\smallskip}
\# meetings & \# participants & Total & Avg per meeting\\
\noalign{\smallskip}\hline\noalign{\smallskip}
38 & 4 & 17h 44min & 28 min\\
\noalign{\smallskip}\hline
\end{tabular}
\end{table}%

This scenario includes 38 recorded meetings with four participants each, whose roles are:
\begin{itemize}
    \item Project manager,
    \item Industrial designer,
    \item User interface designer,
    \item Marketing expert.
\end{itemize}

\begin{figure}[!htb]
\centering
\begin{subfigure}{0.24\linewidth}
\includegraphics[width=0.95\linewidth]{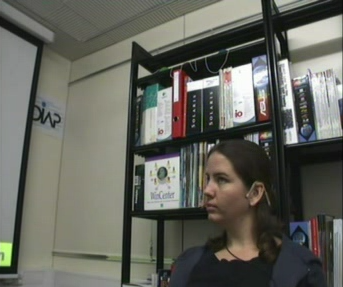}
\end{subfigure}
\begin{subfigure}{0.24\linewidth}
\includegraphics[width=0.95\linewidth]{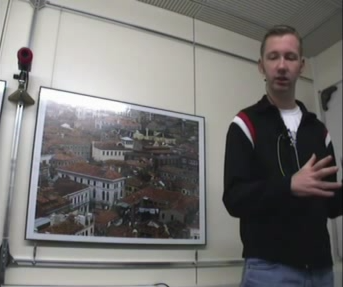}
\end{subfigure}
\begin{subfigure}{0.24\linewidth}
\includegraphics[width=0.95\linewidth]{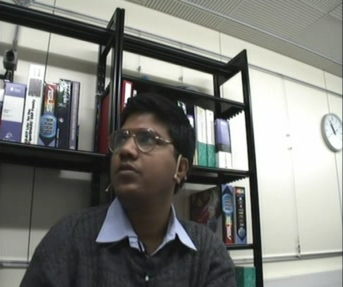}
\end{subfigure}
\begin{subfigure}{0.24\linewidth}
\includegraphics[width=0.95\linewidth]{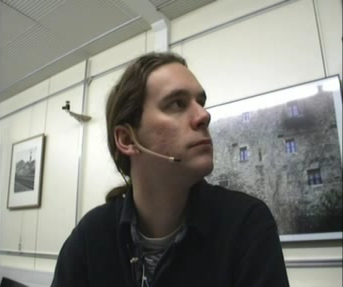}
\end{subfigure}
\caption{Examples of images from the AMI dataset captured by individual cameras.
\label{fig:BDD_train}}
\end{figure}

There are 4 cameras and each one records a single participant as shown in figure~\ref{fig:BDD_train}. 

For the audio part, each participant is recorded with a headset with a microphone. These recordings are then mixed into a single audio file. In the case of our project, only one microphone will be used to record the meetings, so we chose to only use "mix headset" to be in the same application context and to simulate real conditions.

The AMI corpus has detailed ground truth (GT) at the audio level but has no information regarding the video. First, we use this dataset to train and evaluate the performances of the visual-based networks following a cross-validation methodology. Thus, all recorded meetings are divided into five cross-validation folders: four are used to train the models and one for testing.

The CNNs are trained considering individual images only (no general view image) using RGB. The face is detected with \textit{ResNetSSD FaceDetector} from the CAFFE library. 
The samples are grouped in clips of 16 frames successive and labeled as speaker or non-speaker according to the audio-based GT. The images are rescaled to $224 \times 224$ pixels which is the size required for \textit{ResNet}-based network.

\subsection{Evaluation metrics}

To evaluate the different models we proposed and to measure the efficiency of our fusion strategies, we used the macro and micro area under the curve~\cite{Bradley_1997, Hanley_1982} (AUC) applied to the receiver operating characteristic curve (ROC) which is created by plotting the true positive rate (TPR) against the false positive rate (FPR) at various threshold settings. 
The TPR and FPR are defined by:
\begin{align}
    TPR &= \frac{TP}{TP+FN}\label{eq:tpr}\\
    FPR &= \frac{FP}{FP+TN}\label{eq:fpr}
\end{align}

where $TP$ is the number of true positive, $TP+FN$ is the number of real positive cases in the dataset, $FP$ is the number of false positive, $FP+TN$ is the number of real negative cases in the dataset.

ROC is a probability curve and AUC represents the degree or measure of separability. It tells how much the model is capable of distinguishing between classes. Higher the AUC, the better the model will predict the correct class for a new sample.

Macro AUC calculates metrics for each label and computes its unweighted mean. This does not take label imbalance into account. Micro AUC globally calculates metrics by considering each element of the label indicator matrix as a label.

To evaluate diarization performances, we used the Diarization Error Rate (DER), which is the standard metric for evaluating and comparing speaker diarization systems. It is defined as follows:
\begin{equation}\label{eq:der}
    DER = \frac{\textit{False alarm} + \textit{Missed detection} + \textit{Confusion}}{\textit{Total}}
\end{equation}

where \textit{False alarm} is the duration of non-speech incorrectly classified as speech, \textit{Missed detection} is the duration of speech incorrectly classified as non-speech, \textit{Confusion} is the duration of speaker confusion, and \textit{Total} is the total duration of speech in the reference.

\subsection{Assessment of the speaker diarization system}

Concerning the results of \textit{pyBK} on the task of speaker diarization, we did not look after a better performance on this task. Our goal was to figure out if we could improve the result of the active speaker detection task by adding this quickly-obtained information, without a learning phase as it is based on an unsupervised method.

For the \textit{pyBK} speaker diarization system, baseline acoustic features are MFCCs comprising 60 static coefficients computed from windows of 250 ms with a 100 ms shift and with a filter bank of 60 channels. 
The binary key background model (KBM) is determined from a pool of Gaussians, each estimated using windows between 0.5 and 2-second duration set dynamically to ensure a minimum of 1024 components. 
The size of the KBM after Gaussian selection is set to 320. The top number of Gaussians per frame is set to 10. We empirically set the number of initial clusters at 12. 
We used the Jaccard distance metric to select the output clustering solutions, and we employed the elbow method for the selection of the best number of clusters. All parameters were set experimentally.

To assess this speaker diarization system, we used a 5-fold cross-validation on all the audio files of the IDIAP remote control scenario meetings from the AMI corpus. 

As can be seen in Table~\ref{tab:results_pyBK}, the results we have obtained with the parameters described above do not reach those of the state-of-the-art (for example a DER of 12.81\% in~\cite{Dubey_2019}).

\begin{table}[htp]
\small
\centering
\caption{Results obtained with \textit{pyBK} applied on meetings from the IDIAP "remote control scenario".
\label{tab:results_pyBK}}
\begin{tabular}{cccc}
\hline\noalign{\smallskip}
Missed detection rate & False alarm rate & Confusion rate & \textbf{DER}\\
\hline\noalign{\smallskip}
50.52\% & 3.46\% & 16.02\% & 70.0\%\\
\noalign{\smallskip}\hline
\end{tabular}
\end{table}%

Besides, as we can see, the fact that the DER is as high comes mainly from the fact that we have a high rate of missed detection. Indeed, the false alarm rate is only 3.46\%, and the confusion rate is 16,02\%, which is relatively low. These scores mean that even if this approach misses half of the speech segments, the speech segments are generally attributed to the right speakers.

\subsection{Training settings}

Video-net and Audio-net have been trained with a batch size of 20 clips, the Adam optimizer~\cite{Kingma_2015}, and an initial learning rate of 0.05. Training is stopped after 21 epochs.

Besides, since we did not have access to the weights of the Video-net model when we did our experiments, all our models were trained from scratch to have a fair comparison.

\subsection{Evaluation protocol}

\begin{figure}[!htb]
\centering
  \includegraphics[width=\textwidth]{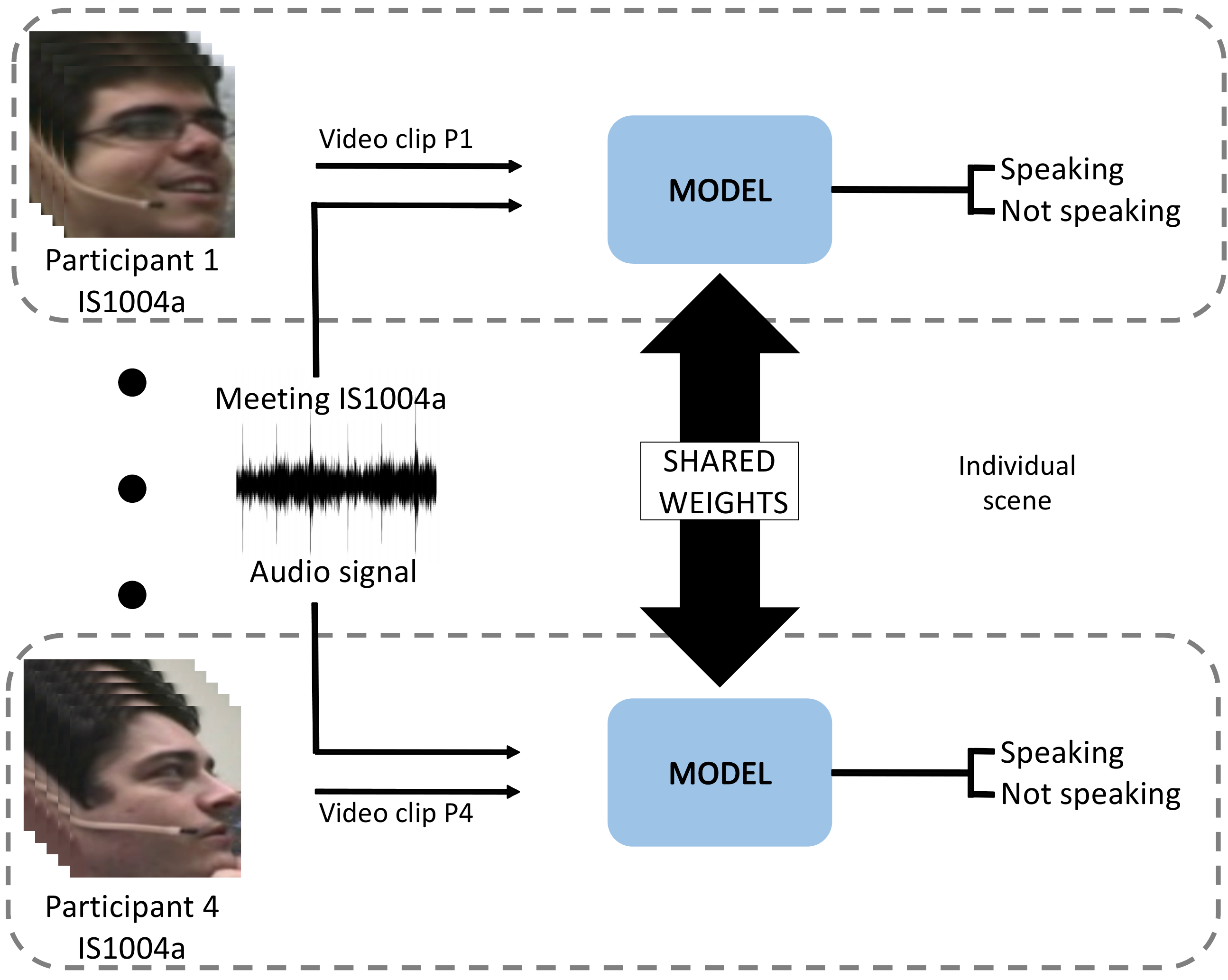}
  \caption{Method used to carry out the evaluation of our different models.
  \label{fig:expe}}
\end{figure}

We present in figure~\ref{fig:expe} how our evaluations were carried out. 

For each meeting, here the IS1004a meeting, we evaluated our system by analysing in parallel the flow of each participant, using the same model (shared weights) for all the participants. 
For the audio part, we use the mix-headset audio recording that consists of voices from all the speakers, so we have a single audio file per meeting.

To summarize, for each evaluated meeting, we have 4 video streams, one for each participant, and a single audio stream, containing the voices of all participants.

\section{Evaluation of our fusion strategies}
\label{sect:evaluation}

In this section, we present the results obtained during our experiments. In table~\ref{tab:resultat_video_audio}, the size of the vector used for fusion is indicated after the name of the neural networks. For example, Video-net\_128 + Audio-net\_16 means that the output of Video-net is a 128 dimensional vector and the output vector of Audio-net is a 16 dimensional vector.

\begin{table}[!htb]
\centering
\caption{Active speaker detection results in terms of Macro AUC using Video-net and Audio-net.
\label{tab:resultat_video_audio}}
\begin{tabular}{ccc}
\hline\noalign{\smallskip}
Fusion & Model & Macro AUC \\
\noalign{\smallskip}\hline\noalign{\smallskip}
& \textBF{Video-net-RGB} & \textBF{72.4\% $\pm$ 3.5}\\
\textbf{Naive} &  & \\
& Video-net-RGB\_128 + Audio-net\_16 & 76.2\% $\pm$ 4.4\\
& \textBF{Video-net-RGB\_128 + Audio-net\_32} & \textBF{76.8\% $\pm$ 4.7}\\
& Video-net-RGB\_128 + Audio-net\_64 & 76.6\% $\pm$ 5.1\\
& Video-net-RGB\_128 + Audio-net\_128 & 76.4\% $\pm$ 5.2\\
\textbf{Attention} & & \\
& Video-ne-RGB\_128 + Audio-net\_16 & 79.4\% $\pm$ 3.7\\
& Video-net-RGB\_128 + Audio-net\_32 & 78.6\% $\pm$ 3.0\\
& \textBF{Video-net-RGB\_128 + Audio-net\_64} & \textBF{79.6\% $\pm$ 2.9}\\
& Video-net-RGB\_128 + Audio-net\_128 & 78.0\% $\pm$ 2.2\\
\noalign{\smallskip}\hline
\end{tabular}
\end{table}

We have not experimented with Audio-net alone because the objective of this network is to support Video-net. By bringing information such as the fact that somebody is speaking and when there is a speaker change. We remind the reader that Audio-net is initially used for speaker recognition and does not allow the detection of the active speaker as it is.

\subsection{Results related to Video-net}

We want to evaluate the potential of visual information in the first step. We started by evaluating the performance of Video-net. As can be seen in the first row of table~\ref{tab:resultat_video_audio}, Video-net obtains an area under the curve equal to 72.4\%. This result will be our baseline for the following experiments.

\subsection{Fusion between Video-net and Audio-net}
\label{sect:fusion_Audio-net_Video-net}

In table~\ref{tab:resultat_video_audio}, we present the results we obtained by merging Video-net and Audio-net. We first present those obtained with the naive fusion, and then the results with the attention-based fusion. 
As can be seen, merging representations slightly improves performance. Indeed, with this fusion, we manage to achieve at best a macro AUC of 76.8\%, an improvement of more than 4\% compared to Video-net alone. 

When we look at attention-based fusion, we can see that there is a very strong improvement in performance. Indeed, for this fusion, we can see an improvement of up to 7.2\% compared to Video-net alone.
The use of attention layers on the output vectors of each modality make fusion more efficient. 

Same evaluations were applied to video, audio, and optical flow.

\subsection{Fusion between audio, video and optical flow}

We present the results of merging video, audio, and optical stream in table~\ref{tab:resultat_video_of_audio}.

\begin{table}[!htb]
\centering
\footnotesize
\caption{Active speaker detection results in terms of Macro AUC using Video-net, Audio-net and optical flow.
\label{tab:resultat_video_of_audio}}
\begin{tabular}{ccc}
\hline\noalign{\smallskip}
Fusion & Model & Macro AUC \\
\noalign{\smallskip}\hline\noalign{\smallskip}
& Video-net-RGB & 72.4\% $\pm$ 3.5\\
& \textBF{Video-net-OF} & \textBF{75.4\% $\pm$ 4.0}\\
\textbf{Naive} &  & \\
& Video-net-RGB\_128 + Video-net-OF\_128 & 81.0\% $\pm$ 4.4\\
& Video-net-RGB\_128 + Video-net-OF\_128 + Audio-net\_16 & 82.2\% $\pm$ 4.0\\
& Video-net-RGB\_128 + Video-net-OF\_128 + Audio-net\_32 & 81.6\% $\pm$ 3.5\\
& \textBF{Video-net-RGB\_128 + Video-net-OF\_128 + Audio-net\_64} & \textBF{83.4\% $\pm$ 4.4}\\
& Video-net-RGB\_128 + Video-net-OF\_128 + Audio-net\_128 & 82.4\% $\pm$ 3.9\\
\textbf{Attention} & & \\
& Video-net-RGB\_128 + Video-net-OF\_128 & 82.4\% $\pm$ 2.3\\
& Video-net-RGB\_128 + Video-net-OF\_128 + Audio-net\_16 & 82.4\% $\pm$ 2.5\\
& \textBF{Video-net-RGB\_128 + Video-net-OF\_128 + Audio-net\_32} & \textBF{84.0\% $\pm$ 2.8}\\
& Video-net-RGB\_128 + Video-net-OF\_128 + Audio-net\_64 & 82.5\% $\pm$ 3.0\\
& Video-net-RGB\_128 + Video-net-OF\_128 + Audio-net\_128 & 83.0\% $\pm$ 3.6\\
\noalign{\smallskip}\hline
\end{tabular}
\end{table}

First of all, we can notice that processing the optical stream alone gives better performance than RGB data processing. Indeed, with optical flow we have a 3\% improvement.

Then if we look at the performances obtained by merging video with optical stream. We can see that this greatly improves performances. In fact, adding this information gives a macro AUC of 81\% for naive fusion and 82.4\% for attention-based fusion. This gives a gain of more than 9\% compared to Video-net alone. 

Let us now focus on naive fusion as shown in table~\ref{tab:resultat_video_of_audio} involving audio modality. We can see that the addition of Audio-net allows a gain of about 4\%. 
However, when looking at the results for attention-based merging, we can see that the performances are similar ($\approx$ 82\%). This is also true for the fusion between Video-net and Audio-net. This fusion allows us to obtain the best performance with a macro AUC of 84\%. However, we can notice that with this fusion we get a smaller standard deviation than with the naive fusion.

\subsection{Alternative and lighter fusion between Video-net and \textit{pyBK}}
\label{sect:fusion_pyBK}

In our approach, we also tried to compare results when using an unsupervised method for speaker diarization based on \textit{pyBK}. The same kinds of experiments were carried out and presented in table~\ref{tab:resultat_video_pybk}.

\begin{table}[!htb]
\centering
\caption{Active speaker detection results in terms of Macro AUC using Video-net, and pyBK's output vectors.
\label{tab:resultat_video_pybk}}
\begin{tabular}{ccc}
\hline\noalign{\smallskip}
Fusion & Model & Macro AUC \\
\noalign{\smallskip}\hline\noalign{\smallskip}
& \textBF{Video-net-RGB} & \textBF{72.4\% $\pm$ 3.5}\\
\textbf{Naive} &  & \\
& \textBF{Video-net-RGB\_128 + pyBK\_16} & \textBF{75.6\% $\pm$ 4.8}\\
& Video-net-RGB\_128 + pyBK\_32 & 74.2\% $\pm$ 5.2\\
& Video-net-RGB\_128 + pyBK\_64 & 74.0\% $\pm$ 3.3\\
& Video-net-RGB\_128 + pyBK\_128 & 74.2\% $\pm$ 4.0\\
\textbf{Attention} & & \\
& Video-net-RGB\_128 + pyBK\_16 & 79.2\% $\pm$ 2.6\\
& \textBF{Video-net-RGB\_128 + pyBK\_32} & \textBF{79.4\% $\pm$ 2.7}\\
& Video-net-RGB\_128 + pyBK\_64 & 78.8\% $\pm$ 3.7\\
& Video-net-RGB\_128 + pyBK\_128 & 78.0\% $\pm$ 2.4\\
\noalign{\smallskip}\hline
\end{tabular}
\end{table}

As we can see, the naive fusion between Video-net-RGB and \textit{pyBK} output vectors gives better performance than Video-net with a gain of about 3\% in the best case. However, this fusion is less efficient than the one between Video-net and Audio-net.
This does not seem significant but we can highlight one possible advantage which concerns the number of parameters involved in computation. This can be of interest if the process has to be embedded.

When we focus on attention-based fusion, we can see that we get the same performances as between Video-net-RGB and Audio-net, an improvement of up to 7\% compared to Video-net alone.

When we focus on the fusion between Video-net, \textit{pyBK}, and optical flow, we can see that the performances are not as good as when we used Audio-net. Indeed, with \textit{pyBK}, the performances are 1\% lower.

\subsection{Discussion}

As we saw earlier, by merging both visual and audio modalities (using \textit{pyBK} or Audio-net) performances were higher than our baseline video-based system. Let us now go a little further in the analysis of our results.

As can be seen in tables~\ref{tab:resultat_video_audio} to \ref{tab:resultat_video_of_pybk}, we show the evolution of the macro AUC as a function of the repartition in the fusion of Video-net-RGB, and Audio-net, or \textit{pyBK}'s output vector. 

\begin{table}[!htb]
\centering
\footnotesize
\caption{Active speaker detection results in terms of Macro AUC using Video-net, pyBK's output vectors, and optical flow.
\label{tab:resultat_video_of_pybk}}
\begin{tabular}{ccc}
\hline\noalign{\smallskip}
Fusion & Model & Macro AUC \\
\noalign{\smallskip}\hline\noalign{\smallskip}
& Video-net-RGB & 72.4\% $\pm$ 3.5\\
& \textBF{Video-net-OF} & \textbf{75.4\% $\pm$ 4.0}\\
\textbf{Naive} &  & \\
& Video-net-RGB\_128 + Video-net-OF\_128 & 81.0\% $\pm$ 4.4\\
& Video-net-RGB\_128 + Video-net-OF\_128 + pyBK\_16 & 81.8\% $\pm$ 3.5\\
& Video-net-RGB\_128 + Video-net-OF\_128 + pyBK\_32 & 81.6\% $\pm$ 3.5\\
& \textBF{Video-net-RGB\_128 + Video-net-OF\_128 + pyBK\_64} & \textBF{82.6\% $\pm$ 2.9}\\
& Video-net-RGB\_128 + Video-net-OF\_128 + pyBK\_128 & 81.4\% $\pm$ 3.7\\
\textbf{Attention} & & \\
& Video-net-RGB\_128 + Video-net-OF\_128 & 82.4\% $\pm$ 2.3\\
& Video-net-RGB\_128 + Video-net-OF\_128 + pyBK\_16 & 82.8\% $\pm$ 2.9\\
& Video-net-RGB\_128 + Video-net-OF\_128 + pyBK\_32 & 81.8\% $\pm$ 3.3\\
& \textBF{Video-net-RGB\_128 + Video-net-OF\_128 + pyBK\_64} & \textBF{83.0\% $\pm$ 2.9}\\
& Video-net-RGB\_128 + Video-net-OF\_128 + pyBK\_128 & 81.2\% $\pm$ 2.9\\
\noalign{\smallskip}\hline
\end{tabular}
\end{table}

The variation of vector size has a small impact on the results. Indeed, the biggest difference one can encounter within the same fusion is only 1.8\% with the Video-net-RGB\_128 + Video-net-OF\_128 + pyBK attention-based fusion.

An interesting point to observe on the overall results is that we did not obtain the best performance when Audio-net has a vector size of 128. This observation is also true when we merge Video-Net with the vector obtained with \textit{pyBK}. Thus, we can deduce that the audio vector is only used here as a support for the video.

Moreover, interesting information to take into account is the number of parameters to be learned. Indeed, \textit{pyBK} is merged with the other networks through a GRU layer. When the output of the GRU layer is set to a vector of size 16, the fusion of \textit{pyBK} with a neural network requires only 66,512 parameters to be trained. 
Table~\ref{tab:param_per_model} shows the number of parameters per model. 

\begin{table}[!htb]
\centering
\caption{Number of parameters per system.
\label{tab:param_per_model}}
\begin{tabular}{cc}
\hline\noalign{\smallskip}
Model & \#parameters \\
\noalign{\smallskip}\hline\noalign{\smallskip}
Video-net-RGB & 33.21M\\
Audio-net\_128 & 21.95M\\
Video-net-RGB\_128 + Audio-net\_16 & 53.56M\\
Video-net\_128 + \textit{pyBK}\_16 & 33.28M\\
\noalign{\smallskip}\hline
\end{tabular}
\end{table}

In this work, we also proposed a lighter alternative by merging Video-Net with the output vector of \textit{pyBK}. Indeed, the Video-net-RGB\_128 + \textit{pyBK}\_16 fusion has 33 million parameters to compute, while Video-net-RGB\_128 + Audio-net\_16 has 53 million. 
Moreover, having fewer parameters means that the learned model will be lighter, for example Video-net-RGB\_128 + \textit{pyBK}\_16 weighs only 133 MB compared to 215 MB for Video-net-RGB\_128 + Audio-net\_16.


In addition to the results presented in the tables~\ref{tab:resultat_video_audio} to \ref{tab:resultat_video_of_pybk}, we have calculated the micro AUC. We have not included these results because it is the same behavior, but it is 1\% lower on all values.

We present, in table~\ref{tab:summurize}, a summary of the best results we have obtained by merging the different modalities and approaches. As can be seen, all the results presented are the result of the attention-based fusion. As we have shown in our experiments, this fusion allows to make the most of each modality and thus to obtain the best detection of the active speaker.

\begin{table}[!htb]
\centering
\footnotesize
\caption{Summary of all our results obtained by merging the different approaches. Only the best fusion result between the different modalities is presented.
\label{tab:summurize}}
\begin{tabular}{ccc}
\hline\noalign{\smallskip}
Fusion & Model & Macro AUC \\
\noalign{\smallskip}\hline\noalign{\smallskip}
& Video-net-RGB & 72.4\% $\pm$ 3.5\\
& Video-net-OF & 75.4\% $\pm$ 4.0\\
\textbf{Attention} & Video-net-RGB\_128 + pyBK\_32 & 79.4\% $\pm$ 2.7\\
\textbf{Attention} & Video-net-RGB\_128 + Audio-net\_64 & 79.6\% $\pm$ 2.9\\
\textbf{Attention} & Video-net-RGB\_128 + Video-net-OF\_128 & 82.4\% $\pm$ 2.3\\
\textbf{Attention} & Video-net-RGB\_128 + Video-net-OF\_128 + pyBK\_64 & 83.0\% $\pm$ 2.9\\
\textbf{Attention} & Video-net-RGB\_128 + Video-net-OF\_128 + Audio-net\_32 & 84.0\% $\pm$ 2.8\\
\noalign{\smallskip}\hline
\end{tabular}
\end{table}

\section{Conclusion}\label{sect:conclusion}

This paper presents a framework for active speaker detection in a meeting context using two fusions between three approaches / modalities. 
During our experiments, we used only one audio representation at a time, either the feature extracted through a CNN (Audio-net) or the feature extracted with \textit{pyBK}, a speaker diarization system. 
The other two representations used are based on visual information. We used the AMI corpus, and from that the video from each camera filming the participants individually of each meeting. We analysed the visual information through 3D CNN (Video-net), which is generally used to classify actions such as "cycling". 
We used this particular neural network to encode the spatiotemporal aspect of a participant in the active speaker detection task. The last feature we used was based on motion via optical flow. This feature is used in combination with video. In order to take this information into account, we made a joint learning with the video with a two-branch network.

We first evaluated Video-net-RGB and used this result as a baseline system. With this experiment, we obtained a macro AUC of 72.4\%.
We then merged Video-net-RGB with Audio-net. We show that merging these two modalities improves results by at least 4\%. We have also noticed by combining these two modalities that the attention-based fusion gives better performances since in the best case, we have a gain of more than 7\%.
Next, we showed that adding motion information with the optical flow greatly increases the results. In fact, merging video with the optical stream allows reaching a macro AUC of 78.8\%, an improvement of more than 6\% compared to Video-net alone. 
Moreover, by adding the audio modality through Audio-net, we show that it further improves performance with a macro AUC of over 83\% regardless of the fusion used.

In a second step, we carried out the same experiments by replacing Audio-net with a lighter alternative which is \textit{pyBK} speaker diarization system. We have shown in the course of our experiments that this alternative allows us to reduce the number of parameters while maintaining performances very close to those obtained with Audio-net.

The continuation of this work consists in using the work presented on the project data with new issues. Indeed, in our project, there may be a higher number of participants. Moreover, the number of participants is not always the same. 
Another difference in the data is that in our case a single microphone is used for all participants. In the case of the AMI corpus data, each participant has a microphone. For this work we used the audio file mixing all the recordings in order to get closer to the recording conditions of our project. 
However, by using a single microphone for all the participants, the distance to the microphone of each participant will be different, so we will have other difficulties. Finally, the video acquisitions of our project are made with a 360 camera. There will therefore be more pre-processing to be done on our data to be in the same conditions as in the AMI corpus.\\

\noindent\textbf{Acknowledgements} This work has been carried out at IRIT and is supported by the LinTO project, funded by Bpi France, as part of the French project Programme d’Investissements d’Avenir 3.

\bibliographystyle{spmpsci}      
\bibliography{bib}   

\end{document}